\pdfoutput=1

\documentclass[11pt]{article}

\usepackage{ACL2023}

\usepackage{times}
\usepackage{latexsym,comment}

\usepackage[T1]{fontenc}

\usepackage[utf8]{inputenc}

\usepackage{microtype}

\usepackage{inconsolata}

\usepackage{natbib}
\usepackage{graphicx}
\usepackage{appendix}
\usepackage{microtype}
\usepackage{bm}
\usepackage{multirow}
\usepackage{enumitem}
\usepackage{soul} 
\usepackage{times}
\usepackage{latexsym}
\usepackage{graphicx}
\usepackage{booktabs}
\usepackage{tcolorbox}
\usepackage{paralist}
\usepackage{enumitem}

\usepackage{mathtools}
\usepackage{amsfonts,amssymb,amsthm}
\usepackage[ruled,noend]{algorithm2e}

\usepackage{cleveref}
\crefformat{section}{\S#2#1#3}
\crefformat{subsection}{\S#2#1#3}
\crefformat{subsubsection}{\S#2#1#3}
\crefrangeformat{section}{\S\S#3#1#4 to~#5#2#6}
\crefmultiformat{section}{\S\S#2#1#3}{ and~#2#1#3}{, #2#1#3}{ and~#2#1#3}

\Crefformat{figure}{#2Figure~#1#3}
\Crefmultiformat{figure}{Figures~#2#1#3}{ and~#2#1#3}{, #2#1#3}{ and~#2#1#3}
\Crefformat{table}{#2Table~#1#3}
\Crefmultiformat{table}{Tables.~#2#1#3}{ and~#2#1#3}{, #2#1#3}{ and~#2#1#3}
\Crefformat{appendix}{#2Appx.~\S#1#3}
\crefformat{algorithm}{Alg.~#2#1#3}
\Crefformat{equation}{#2Eq.~#1#3}

\newcommand{\model}{\textsc{MatCha}}

\usepackage{color, colortbl}
\usepackage{xcolor}
\definecolor{asparagus}{rgb}{0.53, 0.66, 0.42}
\definecolor{applegreen}{rgb}{0.55, 0.71, 0.0}

\newif\ifcomment\commenttrue
\ifcomment
\newcommand{\pinaforecomment}[3]{\colorbox{#1}{\parbox{.8\linewidth}{#2: #3}}}
\else
\newcommand{\pinaforecomment}[3]{}
\fi

\title{\model{} : Enhancing Visual Language Pretraining \\with Math Reasoning and Chart Derendering}
\author{Fangyu Liu$^{\spadesuit\clubsuit}$\thanks{\ \ Work done during Google internship.} \ \ \ Francesco Piccinno$^\clubsuit$ \ \ \ Syrine Krichene$^\clubsuit$ \ \ \ Chenxi Pang$^\clubsuit$ \ \ \ Kenton Lee$^\clubsuit$ \\ \textbf{Mandar Joshi$^\clubsuit$ \ \ \ Yasemin Altun$^\clubsuit$ \ \ \ Nigel Collier$^\spadesuit$ \ \ \ Julian Martin Eisenschlos$^\clubsuit$} \\
$^\clubsuit$Google DeepMind \ \ \ \ $^\spadesuit$University of Cambridge}

\begin{document}

\maketitle

\begin{abstract}
Visual language data such as plots, charts, and infographics are ubiquitous in the human world. However, state-of-the-art vision-language models do not perform well on these data.
We propose \model{} (\textbf{Mat}h reasoning and \textbf{Cha}rt derendering pretraining) to enhance visual language models' capabilities in jointly modeling charts/plots and language data. Specifically we propose several pretraining tasks that cover plot deconstruction and numerical reasoning which are the key capabilities in visual language modeling. 
We perform the \model{} pretraining starting from Pix2Struct, a recently proposed image-to-text visual language model.
On standard benchmarks such as PlotQA and ChartQA, the \model{} model outperforms state-of-the-art methods by as much as nearly 20\%.
We also examine how well the \model{} pretraining transfers to domains such as screenshots, textbook diagrams, and document figures and observe overall improvement, verifying the usefulness of \model{} pretraining on broader visual language tasks.\footnote{Code and models: \href{https://github.com/google-research/google-research/tree/master/deplot}{github.com/google-research/google-research/tree/master/deplot}}\footnote{For questions paper please contact \texttt{fl399@cam.ac.uk} and \texttt{eisenjulian@google.com}.}

\end{abstract}

\section{Introduction}\label{sec:intro}

Visual language is the system that uses tightly integrated textual and visual elements to convey meaning \citep{horn1998visual}.
It is ubiquitous in the human world with typical examples being charts, plots and diagrams existing in places such as textbooks, scientific papers web pages and many more. 
Visual language is also highly complex -- besides texts, its structural units can include line, shape, color, orientation, scale, angle, space, etc. One needs to recognize patterns from these structural units, and perform spatial grouping and/or alignment to extract information for reasoning.

Whilst being prevalent and important, there is little research on visual language understanding from the machine learning community.
Vision-language models pretrained on natural images or image-text pairs crawled from the web perform badly on visual language tasks such as ChartQA \citep{masry-etal-2022-chartqa} and PlotQA \citep{methani2020plotqa} due to the high complexity of jointly modeling language and symbols (more evidence in experiments). Pix2Struct \citep{lee2022pix2struct} is a recently proposed pretraining strategy for visually-situated language that significantly outperforms standard vision-language models, and also a wide range of OCR-based pipeline approaches. Pix2Struct designs a novel masked webpage screenshot parsing task and also a variable-resolution input representation for pretraining an image-to-text encode-decoder Transformer \citep{vaswani2017attention}. In this work, we use Pix2Struct as the base model and further pretrain it with chart derendering and math reasoning tasks.

\begin{figure*}[!ht]
\centering
\includegraphics[width=\linewidth]{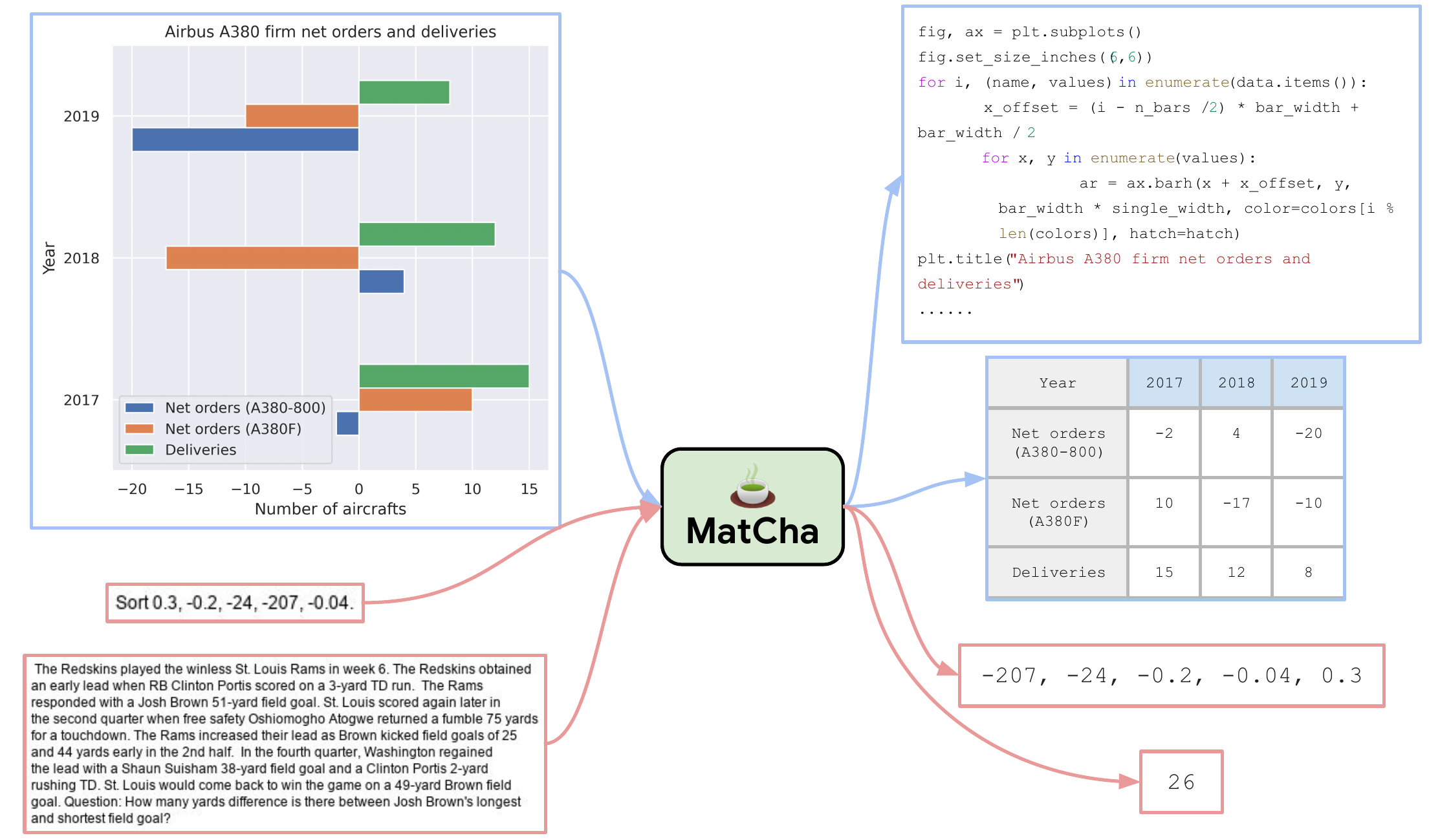}
\caption{\model{} defines two types of pretraining tasks: (1) chart derendering (light blue boxes) and (2) mathematical reasoning (light red boxes). In chart derendering, given a chart, the model needs to decode its underlying rendering code or data table. In math reasoning, given a math question rendered as an image, the model needs to decode its answer. Chart derendering teaches the models layout understanding (including number extraction and their organizations) and math reasoning teaches the models numerical reasoning capabilities. }
\label{fig:front_page}
\end{figure*}

We argue that visual language understanding needs two key ingredients: (1) layout understanding (including number extraction and their organizations) and (2) mathematical reasoning. (1) is required to discover the underlying patterns of the image and organize the elements in the image in a logical form. (2) is needed to operate on the elements extracted from (1) and derive meaningful information demanded by a task or query. Based on these observations, we propose two complementary pretraining tasks for enhancing visual language understanding: \textbf{chart derendering} and \textbf{math reasoning}. In chart derendering, given a plot/chart, the image-to-text model is required to generate its underlying data table or the code used to render it. The second task is math reasoning pretraining. We pick two numerical reasoning dataset MATH \citep{saxton2019analysing} and DROP \citep{dua-etal-2019-drop}, render the input into images and the image-to-text model needs to decode the answers.

We use a suite of visual language tasks to test the effectiveness of our method. Most importantly, we test on ChartQA and PlotQA which are QA datasets about plots and charts. On both datasets, \model{} surpasses even the SOTA model assuming access to charts' underlying data tables and can beat the prior SOTA without gold data table by as much as 20\%. We also test \model{} on chart-to-text summarization tasks and observe clear improvements over Pix2Struct and achieves SOTA on Chart-to-Text \citep{kantharaj-etal-2022-chart} Pew split. Last but not least, to examine if the \model{} pretraining generalizes to datasets beyond the standard plots and charts domain, we also test \model{} on four additional domains where Pix2Struct was evaluated on: documents, illustrations, user
interfaces, and natural images (including datasets, such as textbook QA, Widget Captioning, etc.).
We demonstrate consistent improvement on most additional datasets compared with the base model Pix2Struct. 

To summarize, our contributions are: (1) proposing a set of effective pretraining tasks for visual language learning (2) demonstrating consistent improvements across all evaluated tasks and SOTA results on ChartQA, PlotQA, and Chart-to-Text summarization (Statista set) without accessing the gold data tables; (3) verify that \model{} pretraining transfers to visual language benchmarks beyond the chart \& plot domains and achieve SOTA across a wide range of datasets beyond the chart domain such as textbook VQA and Widget Captioning; (4) comprehensive ablation and analyses to understand the effect of each pretraining component and its impact to downstream performance.

\section{Related Work}\label{sec:rw}

\paragraph{Vision-language research and a lack of attention on visual language.}
Research on vision-and-language has predominately been focusing on natural images. Visually-grounded reasoning datasets such as NLVR2 \citep{suhr-etal-2019-corpus} and MaRVL \citep{liu-etal-2021-visually} are mostly in the natural image domain. Synthesized datasets such as SHAPES \citep{andreas2016neural}, NLVR \citep{suhr-etal-2017-corpus}, and CLEVR \citep{johnson2017clevr} can be seen as in the visual language domain. However, their visual language systems are significantly simpler than those in the real world such as plots and charts. As a result, information extraction from these synthesized datasets is straightforward. Besides, the queries in the synthesized datasets are relatively naive and do not require complex reasoning (e.g., questions can usually be on spatial relations or counting objects). Consequently, current vision-language models can handle the above mentioned synthesized visual reasoning datasets quite well. However, they do not perform well on real-world visual language datasets where both the information extraction and reasoning becomes much more complex (we will show this in \Cref{sec:exp}).

\paragraph{OCR-based \& end-to-end methods for visually-situated language.\footnote{There is a nuanced difference between \emph{visual} language and \emph{visually-situated} language. Most models discussed here are specifically designed for images with significant amount of texts (e.g., documents) and thus they are models primarily for visually-situated language. Visual language data significantly overlaps with visually-situated language data. However, visual language also covers the scenarios where no/few texts are explicitly used but visual objects/patterns are most responsible for presenting information (e.g., certain types of plots).}}
LayoutLM \citep{xu2020layoutlm,huang2022layoutlmv3} leverages a patch-OCR alignment loss to inject an external OCR systems' knowledge into the Transformer model.
PresSTU \citep{kil2022prestu} and PaLI \citep{chen2022pali} also design OCR-aware pretraining objectives where the model needs to predict texts obtained from off-the-shelf OCR systems. 
ChartBERT \citep{akhtar-etal-2023-reading} relies on OCR text and positions to train a transformer encoder.
While OCR systems can be helpful for accurately extracting texts, running them is not cheap. Also, OCR systems do not cover visual language systems that do not explicitly use text. As examples, plots and charts do not always have numbers written explicitly.
In our concurrent work \textsc{DePlot} \citep{liu2022deplot}, we explore combining a chart-to-text translation module (without OCR) with large language models.

Donut \citep{kim2022ocr}, Dessurt \citep{Davis2022EndtoendDR}, and Pix2Struct \citep{lee2022pix2struct} are end-to-end pretrained models for visual language where Donut and Dessurt focus on document understanding and Pix2Struct aim to provide a generic pretrained checkpoint for all visual language tasks. \model's architecture is identical to Pix2Struct -- we continually pretrain a Pix2Struct checkpoint with new objectives.

\paragraph{Learning to reason by designing novel pretraining tasks.} \model{} is related to the literature of designing better pretraining objectives to help the language models (LMs) to reason better since the skill is hard to require through naive language modeling objectives only (e.g, masked language modeling and autoregressive language modeling on raw text).
\citet{geva-etal-2020-injecting, eisenschlos-etal-2020-understanding} generate additional pretraining data focused on (numerical) reasoning through human-written templates. 
\citet{pi-etal-2022-reasoning} synthesize data and programs, and then use program executors to simulate answers. LMs are pretrained to predict the answers given data and programs.
\citet{wu2022insights} explore a wide range of synthetic pretraining tasks and found that even just injecting knowledge as simple as induction and deduction rules could teach LMs to reason. We teach an image-to-text model to reason through mapping charts to data and code, and also directly learning textual math reasoning datasets.

\section{Method}\label{sec:method}

We argue that layout understanding and basic math operation capabilities are the key elements for performing visual language understanding/reasoning. We inject such capabilities to the model by proposing two pretraining tasks: \textbf{chart derendering} (\Cref{sec:plot_derendering}) and \textbf{math reasoning} (\Cref{sec:math_reasoning}) which we introduce in detail in the following sections.

\subsection{Chart Derendering}\label{sec:plot_derendering}
Plots and charts are usually generated by an underlying data table and a piece of code. Code decides the overall layout of the figure (e.g., type, direction, color/shape scheme of the chart) and the underlying data table decides the actual numbers and the groupings of them. Both the data and code are sent to a compiler/rendering engine to create the final image. 
To understand a chart one needs to discover the visual patterns in the image, effectively parse and group them to extract the key information. Reversing the plot rendering process demands all such capabilities and can thus serve as a perfect pretraining task.

In practice, it is challenging to simultaneously obtain charts, their underlying data tables, and their rendering code. To collect sufficient pretraining data, we independently accumulate (chart, code) and (chart, table) pairs. For (chart, code), we crawl all GitHub IPython notebooks with appropriate licenses and extract blocks with figures. A figure and the code block right before it are saved as a (chart, code) pair.\footnote{Note that the code snippet can be noisy since earlier blocks could also be relevant for generating the figure and also the snippet may contain bits of code that is irrelevant to generating the figure. Also note that the data table is frequently missing and usually not hardcoded in the notebook. As a result, we collect (chart, table) pairs separately.} For (chart, table) pairs, we explored two sources. First is to manually write code for converting web-crawled Wikipedia tables from ~\citet{herzig-etal-2020-tapas} to charts.
We randomly combine several plotting options. The key random variables include: using either \texttt{matplotlib} or \texttt{seaborn} as the plotting package; using either bar, line, or pie chart; styles and colors of the charts; whether to show numbers explicitly on the graph; font and size of the texts. Besides our own synthetic data, we also add chart-table pairs generated by 
\citet{methani2020plotqa} (from PlotQA) to diversify the pretraining corpus. The second source is web-crawled chart-table pairs. Websites such as Statista provides both. We directly use the chart-table pairs crawled by \citet{masry-etal-2022-chartqa} (from ChartQA), containing around 20k pairs in total from four websites: Statista, Pew, Our World in Data, and OECD.\footnote{See \Cref{sec:dataset_details} for links.}

Note that to avoid leaking test information for the PlotQA and ChartQA tasks which use the same chart data as pretraining, we only use the chart-table pairs in the training sets for pretraining and test tables/charts are strictly excluded. In ablation study (\Cref{sec:ablations}), we will show that chart-table from both sources are useful and having a diverse set of chart-table pairs is always better. However, using only our synthetic data brings very significant improvement already, suggesting that the concept of chart derendering can be easily transferred to charts of other domains (including real-world charts).


\subsection{Math Reasoning}\label{sec:math_reasoning}
Reasoning over visual language requires (1) effective recognition and grouping of the visual elements and also (2) applying mathematical operations (such as sorting, min/max, averaging, etc.) on top of them. Plot derendering addresses (1) but (2) is still lacking in the current pretraining framework. As a result, we propose to explicitly inject numerical reasoning knowledge to the image-to-text model by learning from textual math datasets.

We use two existing textual math reasoning datasets, MATH \citep{saxton2019analysing} and DROP \citep{dua-etal-2019-drop} for pretraining. MATH is synthetically created, containing two million training examples per module (type) of questions (see \Cref{sec:dataset_details} for a comprehensive listing of modules included in \model{} pretraining).
DROP is a reading-comprehension-style QA dataset where the input is a paragraph context and a question. DROP has 96k question and answer pairs over 6.7K paragraphs.\footnote{Note that for all datasets used for pretraining, we always use only the training set if there exists a split.} To solve questions in DROP, the model needs to read the paragraph, extract relevant numbers and perform numerical computation to predict the answer. We found both datasets to be complementarily helpful. MATH contains large amounts of questions and is categorized which helps us identify math operations needed to explicitly inject to the model. DROP's reading-comprehension format resembles the typical QA format where models need to simultaneously perform information extraction and reasoning. In practice, we render inputs of both datasets into images (concatenating the context and question for DROP). The image-to-text model is trained to decode the answer given the redered image. Examples of MATH and DROP can be found in \Cref{fig:front_page} (in light red).

Besides the two newly proposed pretraining strategies, to prevent catastrophic forgetting, we also keep applying the screenshot parsing pretraining from Pix2Struct \citep{lee2022pix2struct}. Specifically, given screenshot of a website (where parts of the website is masked), the image-to-text transformer needs to predict the underlying simplified HTML code that could render the original unmasked website screenshot. The final pretraining task is a mixture of all aforementioned tasks. We discuss the mixture weights in \Cref{sec:exp_setups}.

\section{Experiment}\label{sec:exp}
We detail our experimental setup in \Cref{sec:exp_setups}, introduce the main results in \Cref{sec:main_results}, and results on additional Pix2Struct tasks in \Cref{sec:p2s_tasks}.

\subsection{Experimental Setups}\label{sec:exp_setups}

\paragraph{Pretraining datasets/tasks.} Overall, we create a mixture of pretraining task that has 40\% of math reasoning, 40\% of chart derendering, and 20\% screenshot parsing. The weight for specific task/dataset is listed in \Cref{tab:mixture_rate}. For chart derendering, we have four sources of data: (1) chart-table pairs synthesized by ourselves, (2) from ChartQA, (3) synthesized in PlotQA, and (4) chart-to-code data. We initially assigned equal weight to the four tasks however noticed training instability since chart-to-code is very hard (the pretraining data is noisy). We thus lower chart-to-code to 4\% and increase all chart-to-table tasks to 12\%. For math reasoning, we assign equal weights to MATH and DROP (both are 20\%).

For pretraining dataset ablation studies, see \Cref{sec:ablations}.

\begin{table}[t!]
    \small
    \centering
  \scalebox{0.85}{
    \begin{tabular}{llcc}
    \toprule
 Component &    Task/Dataset &  Rate & Size  \\
    \midrule
\multirow{2}{*}{\shortstack[c]{Math\\reasoning}}& MATH dataset & 20\% & 2M\\
& DROP & 20\% & 96K \\
\midrule
\multirow{4}{*}{\shortstack[c]{Chart\\derendering}} & Chart-to-code (GitHub; \textbf{ours}) & 4\% & 23M \\
& Chart-to-table (synthetic; \textbf{ours}) & 12\% & 270K \\
& Chart-to-table (ChartQA) &  12\% & 22K \\
& Chart-to-table (PlotQA) & 12\% & 224K \\
\midrule
Pix2Struct & Screenshot parsing &  20\% & 80M \\
\bottomrule
\end{tabular}
}
    \caption{Mixture rates for all tasks in pretraining and the absolute size of each dataset. The mixture rate is used to sample each example within the batch.}
    \label{tab:mixture_rate}
\end{table}

\begin{table}[t!]
    \small
    \centering
  \scalebox{0.85}{
    \begin{tabular}{llcc}
    \toprule
 Task &    Dataset & \# Tables & \# Pairs  \\
    \midrule
\multirow{4}{*}{\shortstack[c]{Chart\\Question\\Answering}} 
& ChartQA (Human) & 4.8K & 9.6K \\
& ChartQA (Machine) & 17.1K & 23.1K \\
& PlotQA (v1) & 224K & 8M \\
& PlotQA (v2) & 224K & 29M \\
\midrule
\multirow{2}{*}{\shortstack[c]{Chart\\Summarization}} & Chart-to-Text (Pew) & 9K & 9K \\
& Chart-to-Text (Statista) & 35K & 35K \\
\bottomrule
\end{tabular}
}
    \caption{Statistics of the finetuning datasets.}
    \vspace{-0.5em}
    \label{tab:finetune_data}
\end{table}

\paragraph{Evaluation datasets.} We evaluate \model{} on multimodal English QA and generation tasks including ChartQA \citep{masry-etal-2022-chartqa}, PlotQA \citep{methani2020plotqa},\footnote{There exists othear chart domain QA datasets such as DVQA \citep{kafle2018dvqa} and FigureQA \citep{kahou2017figureqa}. However, they are both synthetic and SOTA models have already reached $>95\%$ accuracy. We thus focus on more challenging datasets.} and Chart-to-Text summarization \citep{kantharaj-etal-2022-chart}. Both ChartQA and PlotQA are chart domain QA datasets where the input is an image of a chart and a query and the target is an answer string. ChartQA has two subsets: (1) augmented and (2) human where the augmented set is machine generated and thus more extractive and the human set is human written and requires more complex reasoning. PlotQA also has two sets v1 and v2. Similarly, v1 focuses more on extractive questions and v2 requires more numerical reasoning. However, both v1 and v2 are machine generated. Chart-to-Text has two sets as well. They are ``Pew'' and ``Statista'' where the names describe the source of the image examples. For Pew, the gold summaries are automatically extracted from areas around the image. For Statista, the summaries are human written. The sizes of each dataset are described in
~\Cref{tab:finetune_data}.


Beyond chart domain datasets, we additionally evaluate on other datasets used in Pix2Struct \citep{lee2022pix2struct}. We follow the exact same setups and protocols of Pix2Struct by rerunning Pix2Struct experiments but replacing the initial checkpoint with \model{}. See \citet{lee2022pix2struct} for more experimental details. 

\paragraph{Metrics.} For ChartQA and PlotQA, following previous works \citep{masry-etal-2022-chartqa,methani2020plotqa,lee2022pix2struct}, we use relaxed correctness (exact match but tolerating 5\% of numerical error). For Chart-to-Text, we use BLEU4. For all Pix2Struct experiments, we use identical metrics introduced in \citet{lee2022pix2struct}.

\paragraph{Training and inference details.}
We save checkpoint every 200 steps and keep the checkpoint that produces the highest validation score.
Following \citet{lee2022pix2struct}, we finetune models on the ChartQA aug. and human sets together (i.e., one checkpoint for two sets) and use the checkpoint selected on human val set as the final checkpoint for testing.
For PlotQA and Chart-to-Text, we train standalone models for v1, v2, Pew, and Statista sets. For pretraining, we use a batch size of 512 and max sequence length of 192. We pretrain for 100k steps and the final \model{} checkpoint is selected at the 90k step (where the average exact match validation score is the highest).
For downstream tasks finetuning, we use a batch size of 256 and max sequence length of 128.
For ChartQA and Chart-to-Text we finetune for 10k steps and for PlotQA we finetune for 20k steps (since it is significantly larger). Setups for Pix2Struct tasks are the same as the original paper. As for the PaLI baselines, we use the larger 17B variant and finetune for 5k steps and save checkpoints every 1000 steps. All \model{} and Pix2Struct models are pretrained/finetuned with 64 GCP-TPUv3 while PaLI models are finetuned with 128 GCP-TPUv4.

Note that since \model{} is an image-to-text model (without a textual input branch), whenever it is required to input text to the model, the text is rendered as an image. As an example, for QA tasks, we prepend the question as a header above the chart and input the image with question header as a whole to the model.

\subsection{Main Results}\label{sec:main_results}

\begin{table*}[!ht]
    \centering
  \scalebox{0.9}{
    \begin{tabular}{lccccccccccccc}
    \toprule
& \multirow{2}{*}{\shortstack[c]{Gold\\Table?}} & \multicolumn{3}{c}{ChartQA} & & \multicolumn{3}{c}{PlotQA} & &  \multicolumn{3}{c}{Chart-to-Text} & \multirow{2}{*}{\shortstack[c]{avg.\\(all)}}  \\ 
     \cline{3-5}      \cline{7-9}   \cline{11-13}
Model &   &  aug. & human & avg. & & v1 & v2 & avg. & & Pew & Statista & avg. & \\
\midrule
T5 & yes & - & - & 59.8 & & 93.2 & 85.6 & 89.4 & & - & 37.0 & - & -\\
VL-T5 & yes & - & - & 59.1 & & \textbf{96.4} & 84.7 & 90.6 & & - & - & - & -\\
VisionTaPas  & yes & - & - & 61.8 & & 80.2 & 58.3 & 69.3 & & - & - & - & -\\
\midrule 
CRCT & no & - & - & - & & 76.9 & 34.4 & 55.7 & & - & - & - & - \\
VL-T5-OCR  & no & - & - & 41.6 & & 75.9 & 56.0 & 66.0 & & - & - & - & -\\
T5-OCR & no & - & - & 41.0 & & 72.6 & 56.2 & 64.4 & & 10.5 & 35.3 & 22.9 & 42.8 \\
VisionTaPas-OCR & no & - & - & 45.5 & & 65.3 & 42.5 & 53.9 & & - & - & - & - \\
PaLI-17B (res. 224) & no & 11.2  & 15.2 & 13.2 & & 56.9 & 13.1 & 35.0 & & 10.0 & 40.2 & 25.1 & 24.4 \\
PaLI-17B (res. 588) & no & 64.9 & 30.4 & 47.6 & & 64.5 & 15.2 & 39.8 & & 11.2 & \textbf{41.4} & \textbf{26.3} & 37.9 \\
Pix2Struct & no & 81.6 & 30.5 & 56.0 & & 73.2 & 71.9 & 72.5 & & 10.3 & 38.0 & 24.2 & 50.9 \\
\rowcolor{applegreen!20} \model{}  & no &  \textbf{90.2} & \textbf{38.2} & \textbf{64.2} & & 92.3 & \textbf{90.7} & \textbf{91.5} & &  \textbf{12.2} & 39.4 & 25.8 & \textbf{60.5} \\
\bottomrule
\end{tabular}
}
\caption{Main experimental results on two chart QA benchmarks ChartQA \& PlotQA and a chart summarization benchmark Chart-to-Text. Detailed introduction of the baselines can be found in \Cref{sec:baselines}.}
\label{tab:main}
\end{table*}

We summarize the main results in \Cref{tab:main} where we compare \model{} with a wide range of baselines and SOTA models\footnote{For brief introduction of baselines used, please see \Cref{sec:baselines}.} across three chart/plot-domain benchmarks ChartQA, PlotQA, and Chart-to-Text Summarization. On ChartQA, \model{} beats the previous SOTA (without access to the underlying gold data table) Pix2Struct by 8.2\%. Even if we consider models that do assume the existence of gold data tables, they generally underperform \model{} by 3-5\%. The best performing baseline VisionTaPas has a specialized module for modeling tables but still lags behind \model{} by 2.4\%. On PlotQA, \model{} is again the best performing model overall. On the v1 set, VL-T5 with access to underlying data table performs better than \model{} by $\approx4\%$ which is intuitive since PlotQA is a synthetic dataset thus containing relative simple queries and the v1 is the extractive set where queries are even more straightforward. On v2 where questions are related to numerical reasoning, \model{} outperforms all models including the models with access to underlying gold tables.  On Chart-to-Text summarization, \model{} improves upon Pix2Struct on both Pew and Staista and is the new SOTA on Pew. However, \model{} underperforms PaLI-17B (res. 588) on Statista.

Overall, \model{} is clearly the best-performing model with SOTA or competitive performance on every setup and all tasks. 
All baselines without access to gold tables lag behind significantly -- \model{} outperforms the strongest baseline without gold table access Pix2Struct by $\approx10\%$ if we average the performance scores across all datasets.

Among the baselines, we would like to highlight PaLI which is the SOTA for a series of multimodal text-image tasks such as VQA and captioning on natural images and is of a much larger size (i.e., 17B parameters vs. 300M in \textsc{MatCha}). PaLI fails significantly on ChartQA and PlotQA since the challenge in the visual language is distinctly different from that in the natural image domain. Increasing input resolution substantially helps the model's performance (likely due to the better text reading with higher resolution) but this also increases the sequence length (thus also memory and compute) quadratically. PaLI performs reasonably well in Chart-to-Text. We believe this is because the Chart-to-Text task (evaluated by BLEU4) might be more sensitive to textual fluency but less sensitive to factuality as compared with the other two QA tasks. It is expected that PaLI trained with a language modeling objective on natural text will have more advantage under this evaluation setup. 


\subsection{Results on Pix2Struct Tasks}\label{sec:p2s_tasks}

Besides chart/plot domain datasets, we would also like to examine if \model{} transfers to other visual language datasets such as documents, user interfaces, and natural images. We rerun all Pix2Struct finetuning experiments with a \model{} checkpoint and the results are shown in \Cref{tab:matcha_vs_p2s}. On average across all tasks, \model{} outperforms Pix2Struct by 2.3\%. Besides ChartQA, the improvement is also observed in AI2D (QA on textbook diagrams), Widget Captioning (recognizing and captioning widgets in screenshots), DocVQA (QA on scanned documents), etc. Even if we exlucde ChartQA, \model{} can outperform Pix2Struct by 1.6\% on average, suggesting that knowledge learned through \model{} pretraining can be transferred to visual language domains out side of plots/charts.

\begin{table*}[!ht]
    \centering
\setlength{\tabcolsep}{5pt}
  \scalebox{0.9}{
    \begin{tabular}{lccccccccccc}
    \toprule
\multirow{2}{*}{Tasks$\rightarrow$} & \multirow{2}{*}{ChartQA} & \multirow{2}{*}{AI2D} & \multirow{2}{*}{\shortstack[c]{OCR-\\VQA}} & \multirow{2}{*}{RefExp} & \multirow{2}{*}{\shortstack[c]{Widget-\\Cap}} & \multirow{2}{*}{\shortstack[c]{Screen-\\2Words}} & \multirow{2}{*}{\shortstack[c]{Text-\\Caps}} & \multirow{2}{*}{\shortstack[c]{Doc-\\VQA}} & \multirow{2}{*}{\shortstack[c]{Info-\\VQA}} & \multirow{2}{*}{avg.} & \multirow{2}{*}{\shortstack[c]{avg. (excl.\\ChartQA)}} \\
 & & \\
\midrule
Pix2Struct & 56.0 & 40.9 & \textbf{69.4} & 92.2 & 133.1 & \textbf{107.0} & 88.0 & 72.1 & \textbf{38.2} & 77.4 & 80.1 \\
\rowcolor{applegreen!20} \model{} & \textbf{64.2} & \textbf{42.6} & 68.9 & \textbf{94.2} & \textbf{137.7} & 106.2 & \textbf{92.4} & \textbf{74.2} & 37.2 & \textbf{79.7} & \textbf{81.7} \\
\bottomrule
\end{tabular}
}
    \caption{\model{} vs. Pix2Sturct on Pix2Sturct tasks.}
    \label{tab:matcha_vs_p2s}
\end{table*}

\section{Analyses and Discussions}
In this section, we first conduct pretraining ablations in  \Cref{sec:ablations} to understand the usefulness of each pretraining component, then in \Cref{sec:finegrained_analysis} we conduct fine-grained analysis and error analysis to probe \model' strengths and weaknesses.

\subsection{Ablation Study}\label{sec:ablations}

\begin{table}[!ht]
\centering
\setlength{\tabcolsep}{2.2pt}
  \scalebox{0.9}{
    \begin{tabular}{lccccccc}
    \toprule
Setup$\downarrow$ &  aug. &  human &  avg. \\
\midrule
\model{} (full; 50k steps) & 88.6 & \textbf{37.4} & \textbf{63.0} \\
\midrule
\multicolumn{4}{c}{\emph{Component-level ablations}} \\
\midrule
- no math reasoning & 88.2 & 33.0 & 60.6\\
- no chart derendering & 83.7 & 34.4 & 59.1\\
- no Pix2Struct screenshot parsing & 87.8 & 34.9 & 61.4 \\
\midrule
\multicolumn{4}{c}{\emph{Single-task ablations}} \\
\midrule
- no MATH dataset & 88.2 & 36.7 & 62.5 \\
- no DROP dataset & 88.2 & 34.3 & 61.3 \\
- no real-world chart-table pairs & 87.4 & 34.5 & 61.0\\
- no chart-to-code & \textbf{89.1} & 34.6 & 61.9 \\
\bottomrule
\end{tabular}
}
\caption{\model{} pretraining ablations on ChartQA.}
\label{tab:ablations}
\end{table}

We conduct two types of ablations. First, we remove a whole type of pretraining datasets. For example, `no math reasoning' means removing the whole math reasoning component and drops the MATH and DROP datasets. The weights of other datasets in the mixture are proportionally increased. Second, we remove an individual dataset within a component. For example, `no MATH dataset' means removing just MATH dataset but keep other datasets in the math reasoning component untouched. In this scenario, we increase the weight of other math datasets (in this case just DROP) proportionally to maintain the overall weight of the component in the mixture. To reduce compute used, we train one full \model{} model and all its ablated models with 50k steps (the original full \model{} is trained for 100k steps). As a result the \model{} model performance in \Cref{tab:ablations} is slightly lower than the 100k model (63.0 vs. 64.2). The pretrained models are then finetuned and evaluated on ChartQA only.
The full ablation study table is shown in \Cref{tab:ablations} where the first half is component-level ablations and the second half is individual dataset ablation.

\paragraph{The impact of each pretraining component.} On the component-level, we found that removing any major component (math reasoning, chart derendering, and screenshot parsing) would cause a performance drop. The most important component is chart derendering, the removal of which causes a decrease of $\approx 4\%$ averaging across the two sets. Removing math reasoning decreases the avg. score by 2.4\% and removing the continual pretraining of screenshot parsing causes a drop of 1.6\%. We notice that math reasoning is more important to the human set while chart derendering is more important on the augmented set. The findings are likely due to the fact that the human set contains more numerical reasoning questions while the augmented set contains more extractive questions. We also conducted ablations of specific datasets/tasks which we discuss in paragraphs below.


\paragraph{MATH vs. DROP dataset for learning to reasoning.} We have used two datasets, i.e. MATH and DROP, for injecting numerical reasoning capabilities to \model{}. According to \Cref{tab:ablations}, we observe that DROP seems to be more important (the removal of which causes a performance drop of 1.7\% vs. a drop of 0.5\% from the removal of MATH). We conjecture that it is because the reading-comprehension-QA format of DROP is more similar to the downstream task of QA on visual language, where information extraction and reasoning needs to be jointly performed.

\paragraph{Synthetic vs. real-world corpus as pretraining chart-table pairs.} 
We perform another ablation to justify the choice of chart derendering pretraining corpus. Real-world chart-table pairs can increase the diversity and coverage of chart derendering pretraining however we need to explicitly scrape such data from the web. We are interested in understanding to what extent our manually synthesized charts and plots with existing libraries can improve model's performance. The row `no real-world chart-table pairs' shows results of only using synthesized chart-table data by us (i.e., no ChartQA and PlotQA chart-table data). The overall performance drops by 2\%. Interestingly, for the augmented set, the performance only drops 1.2\% but almost 3\% is dropped on the human set. This indicates that extractive questions can usually be solved with synthetic pretraining but the more diverse real-world data (also usually having more sophisticated layout) can benefit reasoning learning more.

\paragraph{The impact of chart-to-code pretraining.} While much of the information in a chart is provided by data table, the code that is used to render the table decides the visual layout (e.g., type of chart and orientation) and attributes (e.g., color) of the data. To test the importance of the chart-to-code pretraining component, we remove it in an ablated pretrained model and the model performance on ChartQA drops by 1.1\% overall. The drop is mainly on the human set where more complex reasoning is required.

\subsection{Fine-grained Analysis and Error Analysis}\label{sec:finegrained_analysis}

\begin{figure}[!ht]
\centering
\includegraphics[width=0.96\linewidth]{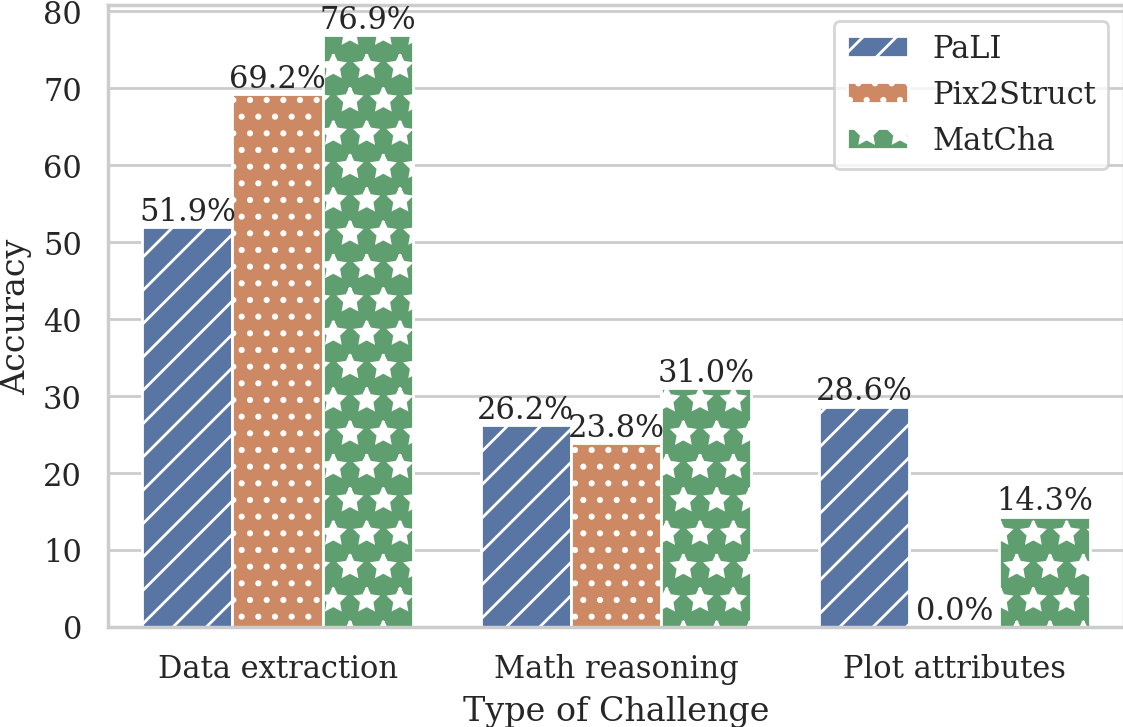}
\caption{Fine-grained by-category performance comparison on ChartQA.}
\label{fig:finegrained}
\end{figure}

\paragraph{Fine-grained analysis.}
To understand the specific aspects of strengths and weaknesses of the models and breakdown the challenges into fine-grained categories, we sample 100 test examples from ChartQA (both augmented and human sets) for further analyses. Specifically, we summarize the challenges of ChartQA into three categories: (1) data extraction (where the model needs to parse a complex query with sophisticated coreference resolution or needs to read numbers when numbers are not explicitly written out), (2) math reasoning (where the model needs to perform one or multiple numerical operations such as min/max/sort/average/etc.), and (3) plot attributes (where the query asks about color/shape/location of specific objects/labels). We manually classify the 100 examples into the three categories and allow an instance to belong to multiple categories when the challenge is multifaceted. After excluding 7 annotation errors, we find 55.9\% questions need complex data extraction, 45.2\% involve math reasoning, and 7.5\% concern plot attributes. We plot the per-category performance of PaLI (res. 588), Pix2Struct and \model{} in \Cref{fig:finegrained}. Overall, all models perform the best on data extraction while math reasoning and plot attributes are more challenging. When compared across models, \model{} improves Pix2Struct in every category and beats PaLI in both data extraction and math reasoning. However, for plot attributes, \model{} lags behind PaLI. This is not significantly reflected in the overall ChartQA performance since plot attribute only concerns less than 10\% of the examples.

\paragraph{Error analysis.} 
 Similar to the fine-grained analysis, we sample 100 errors made by \model{} on ChartQA test set and manually classify the 100 errors into the three categories. After exluding 21 annotation errors, we find 48.3\% of the errors are related to math reasoning, 43.4\% are related to data extraction, and 8.0\% concern plot attributes. We conclude that math reasoning remains to be the major challenge even if \model{} has improved its math reasoning capability compared with Pix2Struct and PaLI. We notice that \model{} still struggles with sophisticated math reasoning questions or numerical computation that requires high precision. An example is shown in Appendix \Cref{tab:hard_numer}.

\paragraph{Case study.} To concretely understand what type of questions \model{} can do better than the baselines, we present several case studies. In \Cref{tab:case_numerical}, we show an example which requires computing average of multiple numbers. Besides \model, PaLI and Pix2Struct's answers are far from the ground truth. In \Cref{tab:case_coref}, we demonstrate an example that requires resolving complex coreference resolution of multiple data points. The model needs to accurately parse the query and find the referenced data points in the chart, then perform a simple numerical computation. \model{} is the only model that gets the correct answer.

\begin{table}[ht!]
  \centering
  \small
  \begin{tabular}{c}
    \begin{minipage}{.45\textwidth}
      \frame{\includegraphics[width=\linewidth, height=75mm]{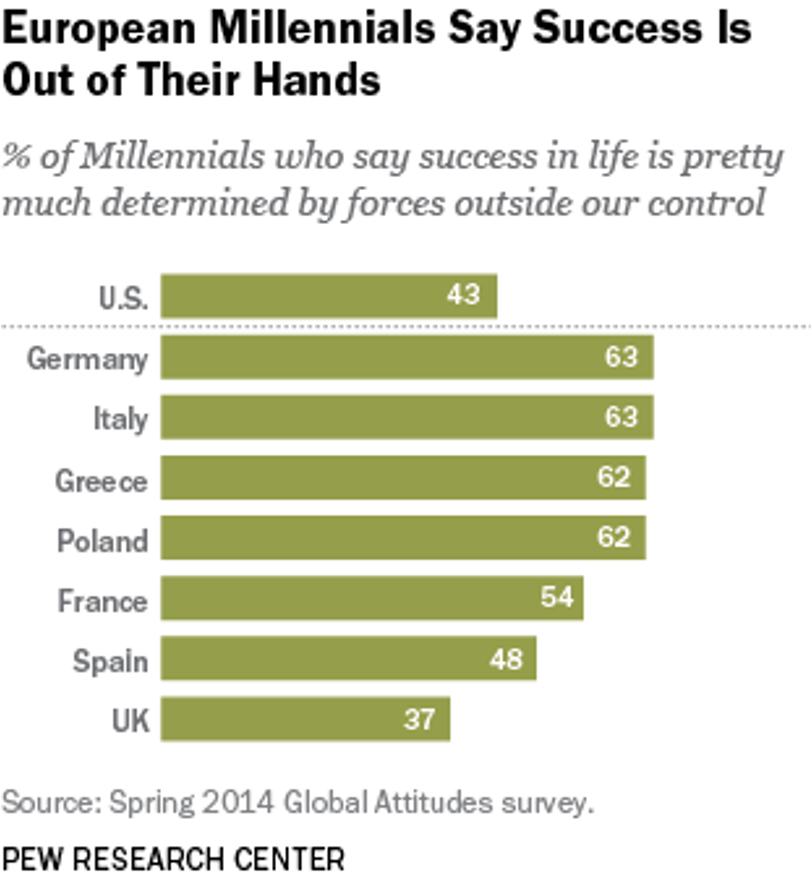}}
    \end{minipage} 
    \\
    \\
    \begin{minipage}[t]{.45\textwidth}
    \texttt{What is the average of last 4 countries' data?} \\
    PaLI: \textcolor{red}{\textbf{40.94}} \ \ \ Pix2Struct: \textcolor{red}{\textbf{40.5}} \ \ \ \model: \textcolor[rgb]{0.4,0.8,0}{\textbf{50.5}}
    \end{minipage}
  \end{tabular}
  \caption{An example that requires strong numerical reasoning skills. \textcolor{red}{\textbf{Red}} and \textcolor[rgb]{0.4,0.8,0}{\textbf{green}} indicate correct and wrong answers respectively.}\label{tab:case_numerical}
\end{table}

\begin{table}[ht!]
  \centering
  \small
  \begin{tabular}{c}
    \begin{minipage}{.48\textwidth}
      \frame{\includegraphics[width=\linewidth]{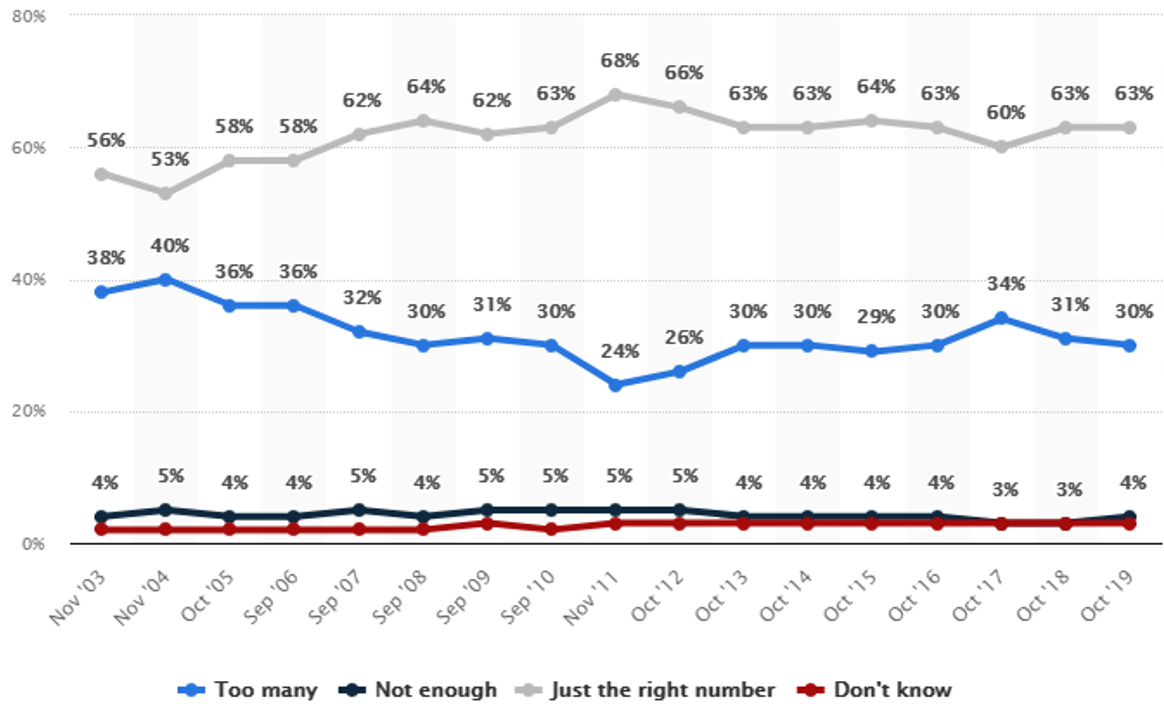}}
    \end{minipage} 
    \\
    \\
    \begin{minipage}[t]{.48\textwidth}
    \texttt{What percentage does 'don't known' and 'just the right number' make up for Oct'17?} \\
    PaLI: \textcolor{red}{\textbf{10}}\ \ \ Pix2Struct: \textcolor{red}{\textbf{21}} \ \ \ \model: \textcolor[rgb]{0.4,0.8,0}{\textbf{63}}
    \end{minipage}
  \end{tabular}
  \caption{An example that requires resolving both coreference resolution and math reasoning.}\label{tab:case_coref}
\end{table}

Besides cases where \model{} succeeded, we also present an example where all models have failed (\Cref{tab:hard_numer}). Questions which require very accurate numerical computation are still very challenging to \model.

\begin{table}[ht!]
  \centering
  \small
  \begin{tabular}{c}
    \begin{minipage}{.48\textwidth}
      \frame{\includegraphics[width=\linewidth]{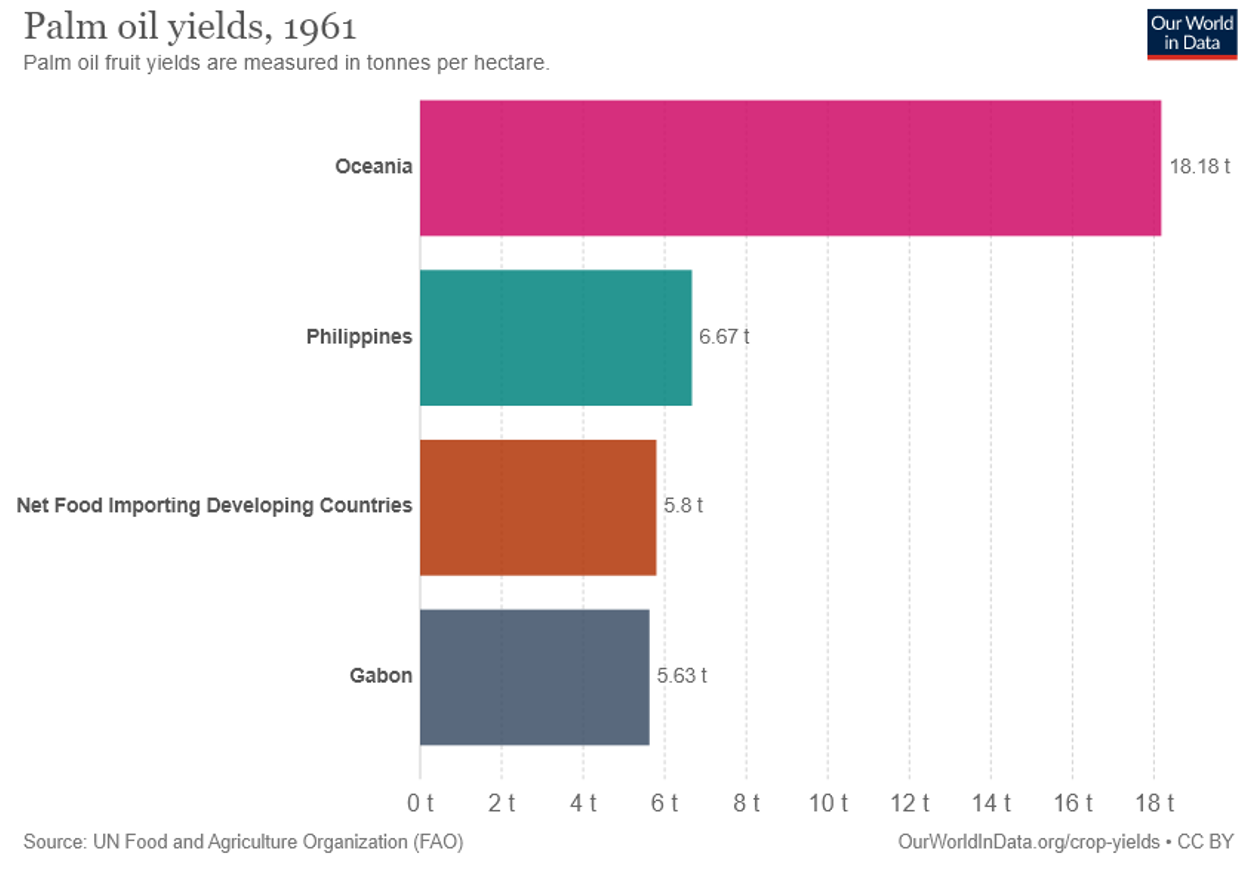}}
    \end{minipage} 
    \\
    \\
    \begin{minipage}[t]{.48\textwidth}
    \texttt{Is the sum of all last three places more than Oceania?} \\
    PaLI: \textcolor{red}{\textbf{Yes}}\ \ \ Pix2Struct: \textcolor{red}{\textbf{Yes}} \ \ \ \model: \textcolor{red}{\textbf{Yes}}
    \end{minipage}
  \end{tabular}
  \caption{An error made by all models including \model{} which requires very accurate numerical computation. The answer should be `No' since 6.67+5.8+5.63=18.1<18.18.}\label{tab:hard_numer}
\end{table}

\paragraph{Continue pretraining Pix2Struct with its original objective.}
It is commonly known that BERT \citep{devlin-etal-2019-bert} is undertrained and simply continuing training BERT with the same objective and on the same data for longer can slightly improve a model's performance \citep{liu2019roberta}. To understand whether such phenomenon persists for \model{} and to what extent does continue pretraining on Pix2Struct screenshot parsing task would improve the model's performance, we continue pretraining Pix2Struct with its original objective and data for 50k steps. We found that continue pretraining indeed improves Pix2Struct's performance (56.0$\rightarrow$57.0 on ChartQA) but is to a much less extent without using the \model{} pretraining components (improving from 56.0 to 64.2).
\section{Conclusion}

We have proposed a pretraining method \model{} for visual language tasks. \model{} injects chart understanding and reasoning knowledge to an image-to-text transformer model by learning to (1) predict the underlying data tables and code given chart images and (2) decode the answers of math questions (rendered in the form of images). \model{} establishes new SOTA on 5 out of 6 setups across three chart domain benchmarks covering both QA and summarization tasks. On visual language tasks beyond the chart domain (e.g., textbook QA and DocVQA), \model{} improves upon Pix2Struct, indicating that the learned knowledge in \model{} pretraining can be transferred outside of the pretraining domain. We conduct comprehensive ablation studies to identify the actual impact of each pretraining component and task and find that chart derendering is essential for extractive questions while math pretraining is important for queries that requires complex reasoning.

\section*{Limitations}\label{sec:limits}
Though we have injected math reasoning skills to \model, error analysis shows that there is still room for improvement on queries requiring complex reasoning. Besides, it remains debatable whether doing math calculation in weight space in a purely end-to-end manner is the most promising path forward.\footnote{See recent works that combine LLMs with calculators \citep{wei2022chain}  or compilers/program executors \citep{cheng2022binding,chen2022program,gao2022pal}.}

Besides math reasoning, \Cref{fig:finegrained} shows that plot attributes is an area where \model{} underperforms PaLI. We conjecture that it is due to \model's lack of massive scale grounded image-text pretraining with rich semantics (which PaLI has using web-scale image-text pairs). While chart-to-code pretraining provides certain level of plot attribute grounding, such plot features are mostly using default options in plotting packages but not explicitly written out in code.

In terms of experimental setup, the reported number is result of a single run. Pretraining is extremely costly especially when there exists more than twenty ablation setups and downstream evaluation tasks. We have collected pretraining and evaluation data points from multiple aspects on various scenarios to verify the robustness of \model. However, we do acknowledge that the paper can benefit from reporting multiple runs given sufficient compute.

Last but not least, it is also worth noting that visual language is an umbrella term. There are other visual language systems beyond the ones discussed in this paper. As an example, comics/manga have their distinct visual lexicon or even grammars \citep{cohn2013visual}.

\section*{Ethics Statement}\label{sec:ethics}
To the best of our knowledge, \model{} has not been trained on sensitive private information and should be of low risk to generate harmful contents. All pretraining and finetuning data are either synthetically created using rules or publicly available data on the web with appropriate permissive licenses.



\bibliography{anthology,custom}
\bibliographystyle{acl_natbib}


\appendix

\section{More Details on Datasets Used}\label{sec:dataset_details}

\paragraph{Chart-table pairs from the web.} The data was originally collected by \citet{masry-etal-2022-chartqa} and came from the below four sources:
\begin{compactitem}
    \item Statista: \url{www.statista.com}
    \item Pew: \url{www.pewresearch.org}
    \item Our World in Data: \url{ourworldindata.org}
    \item OECD: \url{www.oecd.org}
\end{compactitem}

\paragraph{Modules of MATH questions included.} We exclude overly complex math questions and only select the basic modules that would help with numerical reasoning. They are from the two areas of Arithmetic and Comparison. The individual modules included are
\begin{compactitem}
    \item Arithmetic
    \begin{compactitem}
        \item \texttt{add\_or\_sub}
        \item \texttt{add\_sub\_multiple}
        \item \texttt{div}
        \item \texttt{mixed}
        \item \texttt{mul}
        \item \texttt{mul\_div\_multiple}
    \end{compactitem}
     
    \item Comparison
    \begin{compactitem}
        \item \texttt{closest}
        \item \texttt{closest\_composed}
        \item \texttt{kth\_biggest}
        \item \texttt{kth\_biggest\_composed}
        \item \texttt{pair}
        \item \texttt{pair\_composed}
        \item \texttt{sort}
        \item \texttt{sort\_composed}
    \end{compactitem}
\end{compactitem}

Please see \citet{saxton2019analysing} for detailed descriptions about each module and how they are generated.

\section{Details of Baselines}\label{sec:baselines}
We introduce below the details of the baselines used in \Cref{tab:main}.

T5 is an encode-decoder Transformer model proposed by \citet{raffel2020exploring}. The baseline model T5 takes the concatenation of a linearized table (and a query, when the task is QA) as input, and aims to decode the target (answer or summarization). When the gold table is availible, the gold table is used as the input and the chart image is not used directly. VL-T5 proposed by \citet{cho2021unifying} is similar to T5 but also takes a visual input (i.e., the chart image) on the encoder side. VisionTaPas \citep{masry-etal-2022-chartqa} is modified from TaPas \citep{herzig-etal-2020-tapas} to incorporate the visual modality by adding a ViT model \citep{dosovitskiy2021image} and cross-modal fusion layers.
T5-OCR, VL-T5-OCR, and VisionTaPas-OCR are the same model as T5, VL-T5, and VisionTaPas, respectively. However, they do not assume the existence of gold table but use an OCR-based system to extract the data table from the chart image.
The above mentioned models and their performance numbers are all extracted from \citet{masry-etal-2022-chartqa} and \citet{kantharaj-etal-2022-chart}. Please see the original paper for more details.
Classification - Regression Chart Transformer (CRCT) \citep{levy2022classification} is the best performing model on PlotQA according to the PlotQA benchmark on \url{paperswithcode.com}. It uses a detector that extracts all textual and visual elements of chart then processes these elements with a multimodal Transformer.


\end{document}